\documentclass[runningheads]{llncs}

\usepackage{eccv}

\usepackage{eccvabbrv}

\usepackage{graphicx}
\usepackage{booktabs}
\usepackage{multirow}
\usepackage{float}

\usepackage[accsupp]{axessibility}  

\usepackage{hyperref}

\usepackage{orcidlink}

\begin{document}


\title{ReplicateAnyScene: Zero-Shot Video-to-3D Composition via Textual-Visual-Spatial Alignment}


\author{
Mingyu Dong\inst{1}\textsuperscript{*} \and
Chong Xia\inst{1}\textsuperscript{*} \and
Mingyuan Jia\inst{1} \and
Weichen Lyu\inst{1} \and
Long Xu\inst{2} \and
Zheng Zhu\inst{1} \and
Yueqi Duan\inst{1}\textsuperscript{\dag}
    \vspace{1mm}\\
    \textsuperscript{1}Tsinghua University \quad
    \textsuperscript{2}Zhejiang University
    \vspace{1mm}\\
    Project Page: \url{https://xiac20.github.io/ReplicateAnyScene/}
}


\institute{
\inst{1}Tsinghua University \inst{2}Zhejiang University 
}


\institute{}
\maketitle
\begingroup
\renewcommand\thefootnote{} 
\footnotetext{\textsuperscript{*} Equal contribution. \quad \textsuperscript{\dag} Corresponding author.} 
\endgroup

\begin{figure}[H]
  \centering
  \includegraphics[width=0.9\linewidth]{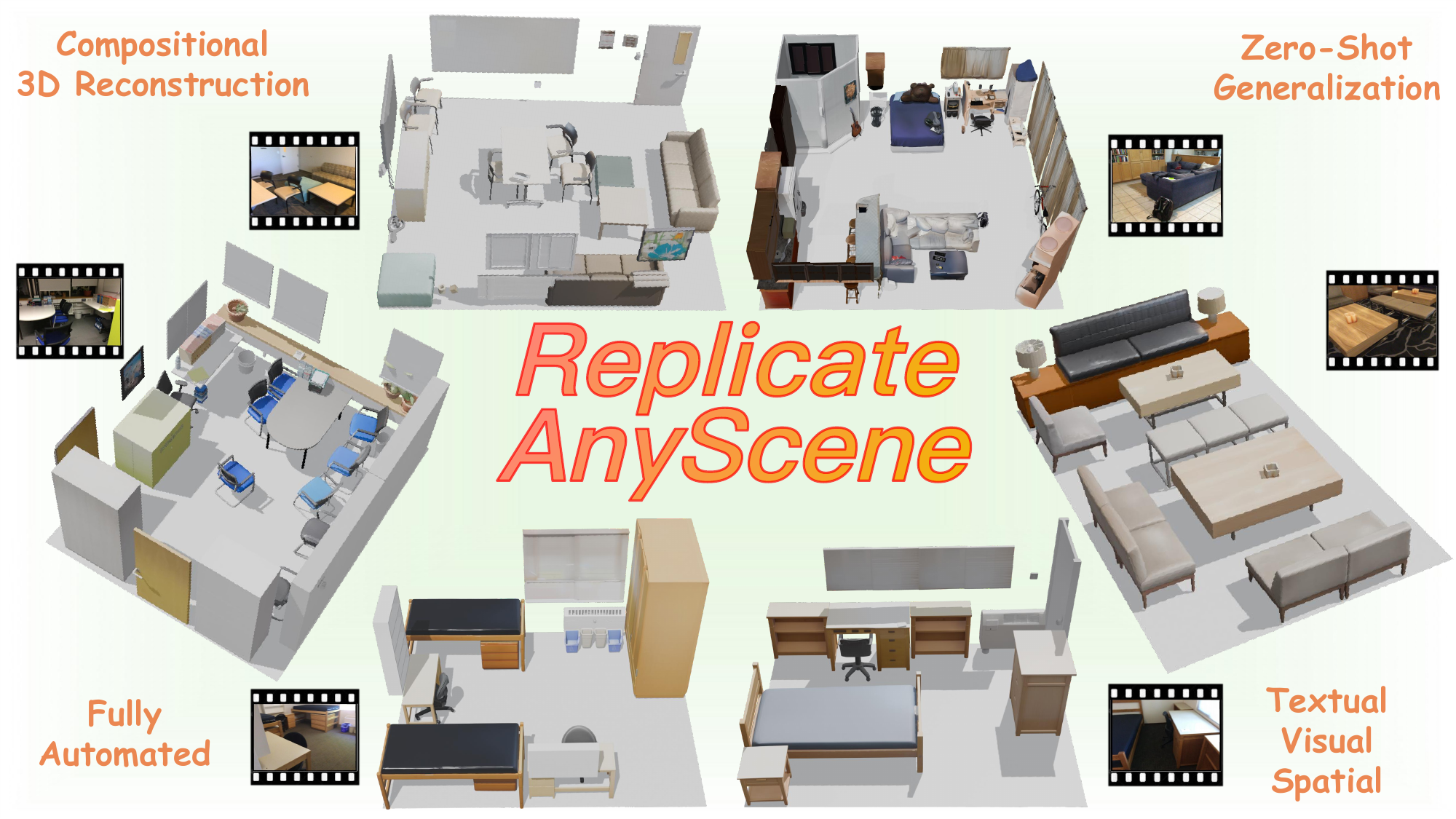}
  \caption{We propose \textbf{ReplicateAnyScene}, a framework for fully automated and zero-shot compositional 3D reconstruction from casually captured videos. Our method extracts and aligns cross-modal priors from vision foundation models to generate semantically coherent and physically plausible 3D scenes.}
  \label{fig:teaser}
\end{figure}

\begin{abstract}
Humans exhibit an innate capacity to rapidly perceive and segment objects from video observations, and even mentally assemble them into structured 3D scenes. Replicating such capability, termed compositional 3D reconstruction, is pivotal for the advancement of Spatial Intelligence and Embodied AI. However, existing methods struggle to achieve practical deployment due to the insufficient integration of cross-modal information, leaving them dependent on manual object prompting, reliant on auxiliary visual inputs, and restricted to overly simplistic scenes by training biases. To address these limitations, we propose \textbf{ReplicateAnyScene}, a framework capable of fully automated and zero-shot transformation of casually captured videos into compositional 3D scenes. Specifically, our pipeline incorporates a five-stage cascade to extract and structurally align generic priors from vision foundation models across textual, visual, and spatial dimensions, grounding them into structured 3D representations and ensuring semantic coherence and physical plausibility of the constructed scenes. To facilitate a more comprehensive evaluation of this task, we further introduce the C3DR benchmark to assess reconstruction quality from diverse aspects. Extensive experiments demonstrate the superiority of our method over existing baselines in generating high-quality compositional 3D scenes.


\end{abstract}

\section{Introduction}
\label{sec:intro}


Compositional 3D scene reconstruction, unlike holistic approaches like NeRF~\cite{mildenhall2021nerf} and 3DGS~\cite{kerbl20233d} that prioritize overall visual fidelity, focuses on extracting complete object geometry and assembling them globally. This makes it inherently more suitable for interactive applications in Spatial Intelligence and Embodied AI. Early compositional 3D scene reconstruction methods~\cite{straub2019replica} typically relied on time-consuming manual annotation and hand-crafted rules. Recently, the rapid development of advanced 3D reconstruction and generation techniques has catalyzed a new paradigm. Consequently, several modern approaches~\cite{yu2025metascenes, ni2025decompositional, yang2025instascene, xia2026simreconsimreadycompositionalscene} have been proposed. These methods typically conceptualize compositional reconstruction as a sequential pipeline: reconstructing 3D scene layouts with semantics, generating independent 3D assets and composing the final scenes.

However, existing methods struggle to achieve robust and automated deployment in real-world scenarios. Their specialized designs are  brittle, manifesting in a heavy dependence on manual intervention, strict requirements for auxiliary depth or semantic inputs, a tendency to generate low-quality objects alongside physically implausible arrangements and a severe restriction to overly simplistic scenes stemming from inherent training data biases. Specifically, DPRecon~\cite{ni2025decompositional} utilizes SDF~\cite{park2019deepsdf} and SDS~\cite{poole2022dreamfusion} for scene decomposition, and necessitates highly accurate depth priors, severely restricting its applicability to simplistic scenes. InstaScene~\cite{yang2025instascene} relies on semantic 3DGS~\cite{kerbl20233d}, which inherently limits its semantic richness. Moreover, lacking explicit relational constraints, it frequently produces object collisions and floating artifacts. Alternatively, MetaScenes~\cite{yu2025metascenes} retrieves assets from complete point clouds, leading to poor consistency and requiring substantial manual annotation. Recently, SimRecon~\cite{xia2026simreconsimreadycompositionalscene} employs view optimization and explicit scene graphs to enhance asset quality and physical plausibility. However, built upon semantic 3DGS, it faces similar bottlenecks, and its view optimization lacks robustness in complex scenarios.

 Driven by the rapid advancement of vision foundation models (VFMs), reliable solutions have emerged for specialized tasks across various dimensions, including VGGT~\cite{wang2025vggt} for geometric reconstruction, SAM3D~\cite{sam3dteam2025sam3d3dfyimages} for 3D asset generation, and VLMs~\cite{Qwen3-VL, lu2024deepseekvl, 2023GPT4VisionSC} for spatial relation reasoning. However, directly assembling these isolated models yields suboptimal results, because they operate independently without cross-dimensional awareness. Their naive integration leads to severe information fragmentation, manifesting as scale ambiguities, identity mismatches, and disjointed spatial layouts. To overcome these bottlenecks, we propose \textbf{ReplicateAnyScene}, a fully automated and zero-shot framework designed to synergize these isolated VFMs. Rather than treating them as simple plug-and-play modules, we progressively ground 2D semantic and perceptual priors into a coherent 3D physical space through uniquely crafted multimodal alignment mechanisms. Our pipeline is structured as a five-stage cascade, enabling the incremental fusion of rich semantic, perceptual, and geometric cues.

Specifically, we begin with \textbf{Progressive Object Discovery} stage, employing a vision-language model (VLM) to scan video frames and extract a non-redundant set of open-vocabulary object categories. Guided by these semantics, the second stage, \textbf{Spatial-Guided Visual Deduplication}, resolves 2D tracking fragmentation by lifting masks into 3D and merging overlapping geometric fragments for consistent instance identities. With unique instances tracked, \textbf{Optimal-View 3D Asset Generation} stage selects the viewpoint maximizing visible surface area to guide the synthesis of high-fidelity 3D meshes. To accurately ground these assets, \textbf{Iterative Visual-Spatial Alignment} stage establishes dense correspondences between rendered and real views, refining object poses through an iterative render-match-optimize cycle. Finally, the last stage, \textbf{Semantic-Aware Scene Refinement}, utilizes VLM-inferred relationships to correct spatial anomalies, such as unnatural tilting or floating, assembling the objects into a physically plausible 3D environment.

Furthermore, to address the lack of standardized evaluation protocols in this task, we introduce the \textbf{C3DR benchmark} to assess compositional reconstruction quality across diverse semantic, visual and geometric dimensions. Extensive experiments demonstrate that our method significantly outperforms existing baselines, setting a new standard for high-quality, fully automated compositional scene generation as shown in \cref{fig:teaser}.

\section{Related Works}

\subsection{3D Indoor Datasets}
Rapid advancements in 3D computer vision have driven the creation of numerous indoor scene datasets~\cite{wald2019rio, baruch2021arkitscenes, yeshwanth2023scannet++, dai2017scannet, chang2017matterport3d, hua2016scenenn}. Typically, these datasets provide scanned meshes from real scenes along with semantic and instance segmentation. However, their holistic representation inherently lacks compositional structure. This fundamental limitation hinders the further advancement of Embodied AI tasks that require object-level manipulation, such as robotic navigation~\cite{chu2026abot, shah2023lm} and interactive tasks~\cite{kim2024openvla, srivastava2022behavior}. To enable more structured and interactive tasks, recent efforts have turned to construct compositional indoor scene datasets. Existing approaches generally rely on human-intensive methods, ranging from manual assembly~\cite{puig2023habitat, ge2024behavior, straub2019replica} to semi-automated pipelines~\cite{avetisyan2019scan2cad, yu2025metascenes}. These automated generation methods procedurally synthesize scenes using rule-based commonsense priors~\cite{deitke2022,paschalidou2021atiss,fu20213dfront} or learned layout priors~\cite{tang2024diffuscene, yang2024holodeck}. Unfortunately, these synthesized environments often suffer from oversimplified layouts and exhibit a significant domain gap from real-world distributions, acting merely as weak augmentations of existing 3D assets. Our approach, by contrast, enables the rapid and automatic generation of compositional indoor scenes directly from monocular video sequences. This eliminates the need for laborious manual labeling while preserving the richness and authenticity of real-world environments, showing great potential for large-scale data generation.

\subsection{Image-Based 3D Asset Generation}
With the rapid advancement of diffusion models~\cite{lipman2022flow,song2020denoising,ho2020denoising} and large-scale 3D asset datasets~\cite{fu20213d, deitke2023objaverse, chang2015shapenet, deitke2023objaversexl}, 3D generation has witnessed significant progress. Early image-conditioned approaches~\cite{xu2024instantmesh, wen2024ouroboros3d, tang2024lgm, long2023wonder3d, liu2023syncdreamer, huang2024mvadapter, yang2024hunyuan3d} relied on multi-view synthesis followed by reconstruction. Recently, methods have shifted toward direct native 3D geometry generation leveraging latent diffusion transformers~\cite{zhao2023michelangelo, zhang2024clay, hunyuan3d2025hunyuan3d, hunyuan3d22025tencent, direct3d, li2025triposg, ye2025hi3dgen,xia2025scenepainter, li2024craftsman, xiang2024structured, kingma2013auto, peebles2023scalable}, sometimes refined by normal maps~\cite{li2024craftsman,ye2025hi3dgen}. Building upon object-level generation, cutting-edge frameworks~\cite{ardelean2025gen3dsr, huang2025midi, meng2025scenegen, yao2025cast, sam3dteam2025sam3d3dfyimages} have begun generating multiple 3D assets and inferring their relative spatial arrangements from a single image. Among these, SAM3D~\cite{sam3dteam2025sam3d3dfyimages} demonstrates remarkable occlusion reasoning and joint geometry recovery capabilities. However, purely single-image scene generation inherently lacks practical application value for complex real-world tasks. The fundamental depth ambiguity of monocular views inevitably leads to inaccurate spatial positioning and distorted holistic scene layouts. Furthermore, naively extending these single-image models to multi-view scenarios introduces critical structural challenges, particularly regarding consistent multi-view conditioning and the lack of robust cross-view positional supervision. Therefore, our framework integrates SAM3D into a unified video-driven pipeline that resolves cross-view ambiguities and extends image-based generation to robust multi-view applications, unlocking its practical value for accurate 3D scene modeling.

\subsection{Compositional 3D Reconstruction}
While early reconstruction techniques~\cite{kerbl20233d, mildenhall2021nerf} predominantly modeled scenes as holistic entities, the growing necessity for object-level interaction has spurred the development of compositional paradigms~\cite{ni2025decompositional, yang2025instascene, yu2025metascenes,xia2026simreconsimreadycompositionalscene}. Despite their potential, current methodologies exhibit distinct architectural limitations. For instance, DPRecon~\cite{ni2025decompositional} formulates scene decomposition through SDF~\cite{park2019deepsdf} and SDS~\cite{poole2022dreamfusion}. However, its reliance on rigorous normal and depth priors, coupled with extensive optimization times, confines its scalability to simplistic topologies. Adopting a dense reconstruction strategy, InstaScene~\cite{yang2025instascene} performs instance segmentation on 3DGS~\cite{kerbl20233d} representations followed by generative completion. This workflow not only mandates redundant multi-view captures and manual category prompts but also lacks structural positions, inevitably resulting in inter-object penetrations and floating artifacts. To circumvent geometry optimization, MetaScenes~\cite{yu2025metascenes} introduces a retrieval-augmented synthesis approach using intact point clouds. Yet, retrieving disjointed assets often disrupts global spatial coherence and demands heavy human supervision for pose alignment. More recently, SimRecon~\cite{xia2026simreconsimreadycompositionalscene} integrates scene-graph relations and active view optimization to enhance scene quaility. Nonetheless, it inherits the stringent multi-view input constraints and suffers from unstable view optimization rendering in complex scenarios. To break free from these constraints, our framework progressively align multimodal priors from diverse VFMs, achieving a purely zero-shot and fully automated compositional reconstruction directly from casual videos with outstanding performance.
\section{Method}
\begin{figure}[tb]
  \centering
  \includegraphics[width=\linewidth]{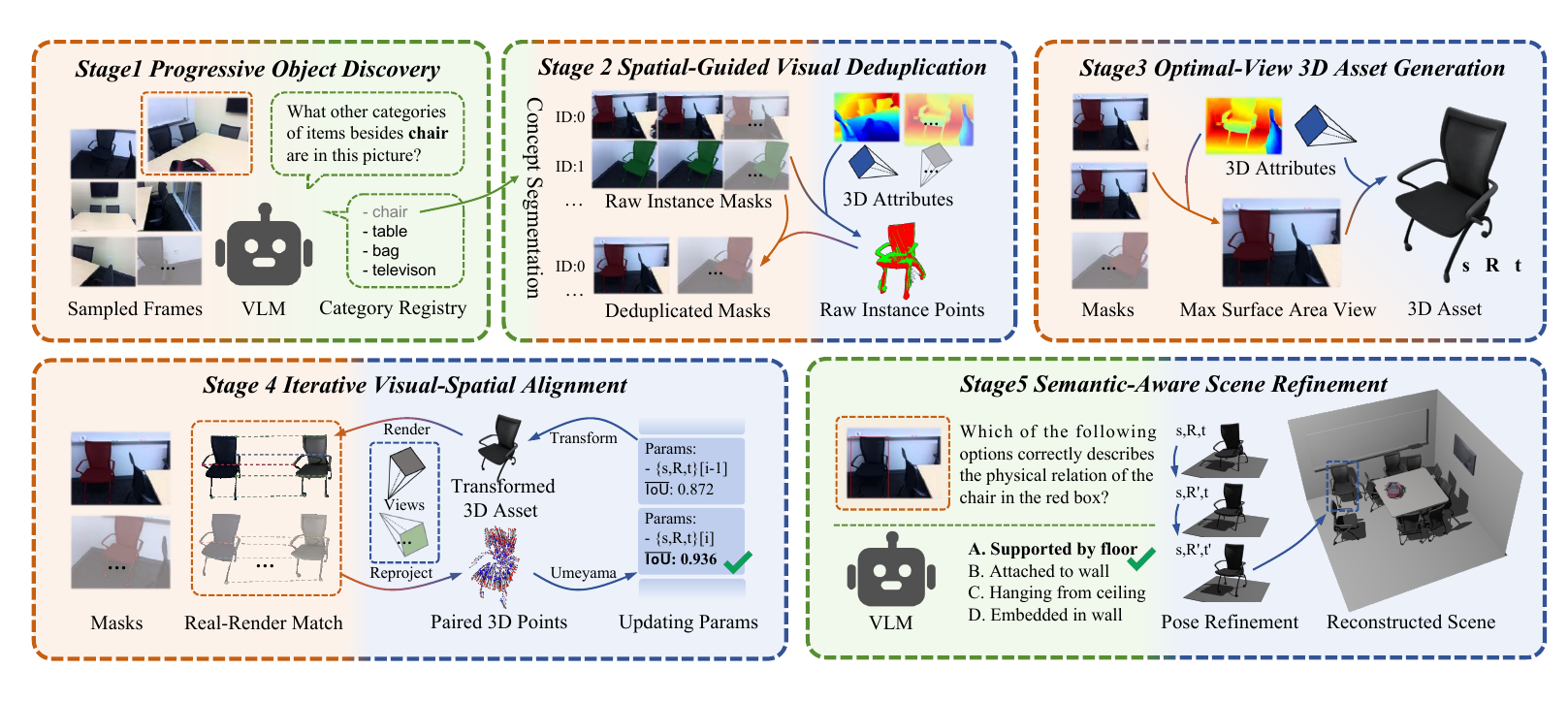}
    \caption{\textbf{Framework Overview.} Our pipeline consists of a five-stage cascade where each stage is specifically designed to resolve targeted alignment gaps among our three core modalities including \textbf{textual (green), visual (orange), and spatial (blue)}. The gradient backgrounds and multi-colored dashed borders within each module explicitly illustrate the specific cross-modal alignment process occurring at that step.}
  \label{fig:pipeline}
\end{figure}
In this section, we present the ReplicateAnyScene framework. As illustrated in \cref{fig:pipeline}, we decompose this complex task into a sequential five-stage pipeline. To effectively harness the diverse capabilities of vision foundation models, each stage is designed to bridge a specific modality gap among textual, visual, and spatial information. This systematic alignment ensures the seamless integration of semantic, visual and geometric priors, ultimately yielding a structurally coherent and physically plausible 3D environment.


\subsection{Progressive Object Discovery}
The primary objective of this stage is to automatically construct a comprehensive and non-redundant open-vocabulary category list for the input video. Naively feeding an entire video sequence into a vision-language model (VLM) overwhelms its contextual capacity, inevitably leading to severe object omission. To circumvent this bottleneck, we propose a progressive discovery strategy. First, to balance computational efficiency with geometric completeness, we select $N_k = 20$ frames maximizing spatial coverage via the space-aware frame sampling algorithm~\cite{wu2025spatialmllmboostingmllmcapabilities}. These frames are then processed sequentially by a VLM~\cite{Qwen3-VL} paired with a dynamic category registry. For each frame, the VLM is prompted to identify only novel objects semantically distinct from the current registry (e.g., treating ``sofa'' and ``couch'' as identical). This iterative querying deduplicates synonymous concepts and yields a highly concise object label set.

\subsection{Spatial-Guided Visual Deduplication}
Given the comprehensive category list derived from the previous stage, the subsequent objective is to establish consistent instance-level segmentations for these target objects across the video sequence. While we can leverage SAM3's~\cite{carion2025sam3segmentconcepts} concept segmentation and temporal tracking capabilities using the identified text queries, relying solely on 2D visual tracking inevitably leads to identity inconsistency. Specifically, discontinuous observations caused by object re-entry or long-term occlusions frequently fragment a single physical instance into multiple redundant mask sequences.

Recognizing that a physical object possesses unique and consistent spatial coordinates, we introduce a spatial-guided deduplication mechanism. By lifting the 2D masks into a unified 3D space, we utilize geometric proximity as the definitive criterion to robustly merge disjointed tracking fragments.

First, we extract per-frame depth maps and camera poses using VGGT~\cite{wang2025vggt}. For each fragmented instance track generated by SAM3, we back-project its associated 2D masks into 3D space, aggregating them into a unified instance-level point cloud. To consolidate identities, we evaluate the geometric overlap exclusively between point clouds that share the same semantic category. 

Specifically, we calculate the average nearest-neighbor distance $r_i$ to represent the density of each point cloud $P_i$:
\begin{equation}
r_i = \frac{1}{|P_i|} \sum_{p \in P_i} d_{\text{nn}}(p),
\end{equation}
where $d_{\text{nn}}(p)$ denotes the Euclidean distance from point $p$ to its closest neighbor in $P_i$. Defining the proximity threshold as $\tau_i = 3 r_i$, we formulate the overlap ratio of $P_i$ relative to $P_j$ as:
\begin{equation}
\text{ov}_{i \to j} = \frac{1}{|P_i|} \sum_{p \in P_i} \mathbb{I}\big(d(p, P_j) < \tau_i\big),
\end{equation}
where $d(p, P_j) = \min_{q \in P_j} \|p - q\|_2$ is the minimum distance from $p$ to $P_j$. The reverse overlap ratio $\text{ov}_{j \to i}$ is computed symmetrically. If either $\text{ov}_{i \to j} > 0.5$ or $\text{ov}_{j \to i} > 0.5$, we classify $P_i$ and $P_j$ as the same physical entity. Ultimately, a Union-Find data structure propagates these pairwise matches to assign globally consistent instance identifiers.

\subsection{Optimal-View 3D Asset Generation}
To instantiate high-fidelity 3D geometries for the consolidated instances, we employ SAM3D~\cite{sam3dteam2025sam3d3dfyimages}, which synthesizes a textured mesh conditioned on a single 2D mask and its corresponding 3D point map. Since each object is observed across multiple frames, selecting the optimal viewpoint is critical. Relying simply on the largest 2D mask area is often misleading due to perspective distortions and camera proximity. Consequently, our objective is to comprehensively maximize the actual 3D spatial information captured in the selected views.

To formulate this selection mathematically, consider a set of candidate masks $\{M_k\}_{k=1}^K$ for a given instance. Ideally, we aim to choose the optimal view $k^*$ that maximizes the mutual information between the complete, unobserved underlying surface $\mathcal{S}$ and the partial surface $\hat{\mathcal{S}}_k$ observed from the $k$-th viewpoint:
\begin{equation}
    k^* = \arg\max_{k} \mathcal{I}(\mathcal{S}; \hat{\mathcal{S}}_k).
\end{equation}

Since computing this theoretical mutual information directly is intractable, we propose a robust geometric proxy. Positing that the density of recoverable geometric detail strongly correlates with the visible surface extent, we utilize the lifted 3D surface area as our criterion. Specifically, for each candidate mask $M_k$, we lift its valid pixels into 3D using the corresponding depth map, construct a local triangular mesh, and compute its total surface area:
\begin{equation}
    k^* = \mathop{\arg\max}\limits_{k \in \{1,\dots,K\}} \mathrm{Area}(\hat{\mathcal{S}}_k)= \mathop{\arg\max}\limits_{k \in \{1,\dots,K\}} \sum_{T \in \mathcal{T}(M_k)} \mathrm{Area}(T),
\end{equation}
where $\mathcal{T}(M_k)$ denotes the set of triangles in the mesh derived from mask $M_k$. Ultimately, we select the $k^*$-th view as the optimal observation and subsequently feed its corresponding mask $M_{k^*}$ and the associated point map into SAM3D to generate the final 3D object asset.

\subsection{Iterative Visual-Spatial Alignment}
Although SAM3D~\cite{sam3dteam2025sam3d3dfyimages} provides the initial transformation parameters $T^{(0)} = \{s^{(0)}, R^{(0)}, t^{(0)}\}$ for an object, the resulting 3D position often exhibits significant spatial misalignment with the global scene structure. To achieve a precise alignment, we propose a render-match-optimize iterative alignment algorithm inspired by the classical Iterative Closest Point (ICP) framework~\cite{121791}. However, unlike traditional ICP which relies purely on geometric proximity and is notoriously prone to local optima under poor initialization, our method is fundamentally driven by visual correspondences. By leveraging multi-frame perceptual information, we establish explicit point associations based on rich texture and semantic cues, effectively achieving spatial alignment.

\noindent \textbf{Initialization.}
Let $k^*$ be the optimal view computed in the previous stage, and define its local temporal neighborhood as $\mathcal{V} = \{k^* - r, \dots, k^* + r\}$. For each $v \in \mathcal{V}$, we denote the observed RGB image and estimated depth map as $I_{\text{real},v}$ and $D_{\text{real},v}$, respectively.

\noindent \textbf{Iterative Optimization.}
In the $i$-th iteration ($i = 1, \dots, K$), we refine the current transformation estimate $T^{(i-1)}$ as follows:

\textbf{Step 1: Rendering and Matching.}
For each view $v \in \mathcal{V}$, we render the object using the current transformation $T^{(i-1)}$ to obtain the rendered RGB image $I^{(i)}_{\text{ren},v}$ and depth map $D^{(i)}_{\text{ren},v}$. We employ MASt3R~\cite{mast3r_eccv24} (denoted as $\Phi$) to establish dense 2D correspondences between the real and rendered images. This visual matching process provides strong perceptual constraints, bypassing the geometric nearest-neighbor matching errors common in standard ICP:
\begin{equation}
    \mathcal{C}^{(i)}_v = \{(p_j, q_j)\} = \Phi(I_{\text{real},v}, I^{(i)}_{\text{ren},v}),
\end{equation}
where $p_j$ and $q_j$ are corresponding pixel coordinates in the real and rendered images, respectively.

\textbf{Step 2: 3D Lifting and Aggregation.}
We lift these 2D matches into 3D world space using camera intrinsics $K$ and extrinsics $T_v = [R_v \mid t_v]$. Let $\pi^{-1}(\cdot; K, T_v)$ denote the back-projection function that maps a pixel and its depth to a 3D point. Aggregating these lifted points across multiple temporal views introduces a robust, multi-perspective constraint:
\begin{align}
    \mathcal{P}_{\text{real}} &= \bigcup_{v \in \mathcal{V}} \bigcup_{j} \pi^{-1}\big(p_j, D_{\text{real},v}(p_j); K, T_v\big), \\
    \mathcal{P}_{\text{ren}}  &= \bigcup_{v \in \mathcal{V}} \bigcup_{j} \pi^{-1}\big(q_j, D^{(i)}_{\text{ren},v}(q_j); K, T_v\big).
\end{align}

\textbf{Step 3: Similarity Alignment.}
We compute an updated similarity transformation $T^{(i)} = \{s^{(i)}, R^{(i)}, t^{(i)}\}$ that best aligns $\mathcal{P}_{\text{ren}}$ to $\mathcal{P}_{\text{real}}$ by minimizing the geometric error over all matched point pairs. Using the Umeyama algorithm~\cite{88573}, we solve:
\begin{equation}
    T^{(i)} = \arg\min_{s, R, t} \sum_{(p, q) \in \mathcal{M}^{(i)}} \| p - (s R q + t) \|_2^2,
\end{equation}
where $\mathcal{M}^{(i)}$ denotes the set of 3D point correspondences derived from $\{(p_j, q_j)\}$.

\noindent \textbf{Selection.}
After $K$ iterations, we select the optimal transformation by evaluating the mean Intersection-over-Union (IoU) between rendered and real masks across all views in $\mathcal{V}$. Specifically, let $M^{(i)}_{\text{ren},v}$ denote the rendered mask for view $v$ at iteration $i$, and $M_{\text{real},v}$ denote the corresponding real mask. The final parameters are chosen as:
\begin{equation}
    T^* = T^{(i^*)}, \quad \text{where} \quad i^* = \mathop{\arg\max}\limits_{i \in \{1, \dots, K\}} \frac{1}{|\mathcal{V}|} \sum_{v \in \mathcal{V}} \mathrm{IoU}\big(M^{(i)}_{\text{ren},v}, M_{\text{real},v}\big).
\end{equation}



\subsection{Semantic-Aware Scene Refinement}
While the previous iterative alignment minimizes geometric projection errors, purely visual optimization often fails to enforce strict physical constraints. This frequently results in artifacts such as unnatural tilting, floating above surfaces, or interpenetration. To rectify this, we introduce a semantic-aware refinement mechanism driven by a vision-language model (VLM)~\cite{Qwen3-VL}. The VLM infers spatial relationships among objects, providing semantic priors to guide deterministic geometric corrections via predefined rule sets tailored to various relationships. To execute these rules, we strategically exploit the canonical coordinate space of the generated 3D assets, utilizing their intrinsically aligned orthogonal axes as reliable anchors for precise pose manipulation. 

For example, given a ``supported by'' relationship, we apply its corresponding rule: we first refine the rotation matrix $R$ to align the object's canonical vertical axis with global gravity, naturally preserving its semantic forward-facing axis. Subsequently, we adjust the translation vector $t$ vertically until the object's bottom surface strictly contacts the supporting plane. Systematically applying these relationship-specific rules eliminates residual spatial violations, ultimately yielding a physically plausible 3D scene.

\section{Experiments}
In this section, we comprehensively evaluate the proposed ReplicateAnyScene framework. We first introduce a newly curated Compositional 3D Reconstruction Benchmark (C3DR) to rigorously assess scene generation across textual, visual, and spatial dimensions. We then provide extensive quantitative and qualitative comparisons against state-of-the-art baselines, demonstrating our framework's superiority in producing semantically accurate and physically plausible 3D environments. Finally, we conduct detailed ablation studies to validate the effectiveness and structural necessity of each core stage within our pipeline.

\begin{figure}[tb]
  \centering
  \includegraphics[width=\linewidth]{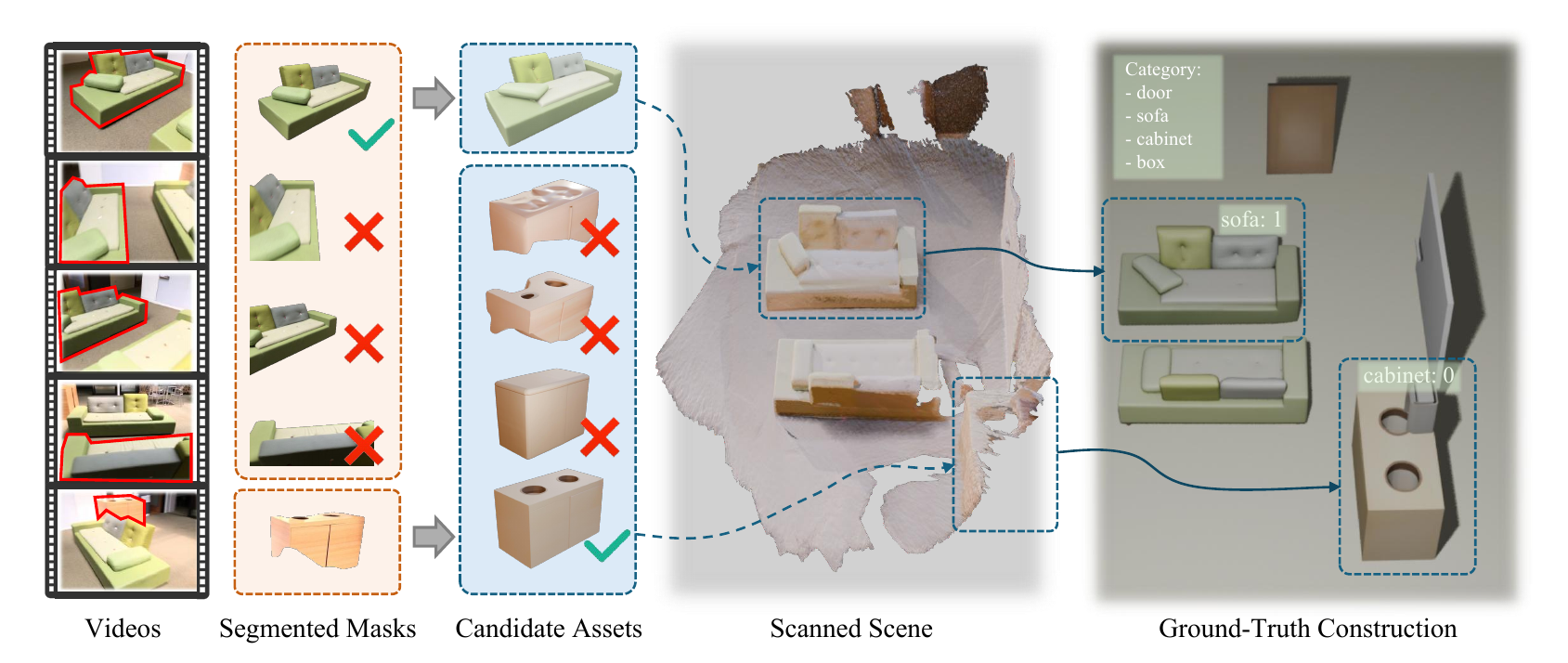}
  \caption{\textbf{Overview of the proposed C3DR benchmark.} This figure illustrates the ground-truth construction pipeline. We first manually select the semantic text and corresponding video frames for each object. We then independently condition a diverse suite of generation models on multiple individual input views for the final selection of high-quality 3D assets. Finally, professional modelers manually place these assets against reference scene meshes to ensure accurate spatial layouts.}
  \label{fig:benchmark}
\end{figure}

\subsection{C3DR Benchmark}
Despite recent progress in compositional 3D reconstruction, the community lacks a standardized benchmark evaluating methods across textual, visual, and spatial dimensions. To bridge this gap, we introduce the Compositional 3D Reconstruction Benchmark (C3DR). C3DR comprises 50 diverse and highly complex scenes carefully curated from Replica~\cite{straub2019replica}, ScanNet~\cite{dai2017scannet}, ScanNet++~\cite{yeshwanth2023scannet++}, and real-world videos from YouTube. We specifically preselect environments featuring intricate object interactions and complex spatial relationships to rigorously evaluate the limits of current scene reconstruction paradigms.

The ground-truth construction operates across three fundamental levels including textual semantics, object geometry, and spatial layouts. For textual annotations, we leverage the native semantic labels from established datasets and manually complete the annotations for small or previously overlooked items. For real-world videos, we provide comprehensive manual semantic tagging. To acquire high-fidelity 3D object assets, we independently condition a diverse suite of generation models on multiple individual input views. This suite includes proprietary models like Rodin~\cite{zhang2024clay} alongside publicly available models such as SAM3D~\cite{sam3dteam2025sam3d3dfyimages}, Trellis~\cite{xiang2024structured}, and Hunyuan3D~\cite{hunyuan3d2025hunyuan3d}. For objects suffering from severe occlusion, we apply generative completion~\cite{nanobanana2026} and super-resolution models~\cite{rombach2022high} to restore missing details. We then manually select the highest-quality 3D asset from these diverse candidates. Finally, professional 3D modelers manually align and place these selected assets against the reference scene meshes. This meticulous manual placement ensures physically accurate spatial locations and structural consistency in the final compositional layout. \cref{fig:benchmark} illustrates the construction pipeline and a representative example of our benchmark.

\subsection{Experimental Settings}
\noindent \textbf{Benchmark.} We conduct experiments on 50 diverse scenes from our C3DR benchmark using only raw RGB video frames as input.

\noindent \textbf{Baselines.} We mainly compare our approach against the state-of-the-art baseline SimRecon~\cite{xia2026simreconsimreadycompositionalscene} and the retrieval-based paradigm MetaScenes~\cite{yu2025metascenes}. DPRecon~\cite{ni2025decompositional} is excluded due to its strict reliance on precise normal and depth annotations and a prohibitive processing time exceeding 10 hours per scene. Additionally, We also omit InstaScene~\cite{yang2025instascene} since its core generation module remains closed-source. As SimRecon demonstrates latest superior performance, we select it as the representative for such optimization-based approaches.

\begin{figure}[t]
  \centering
  \includegraphics[width=\linewidth]{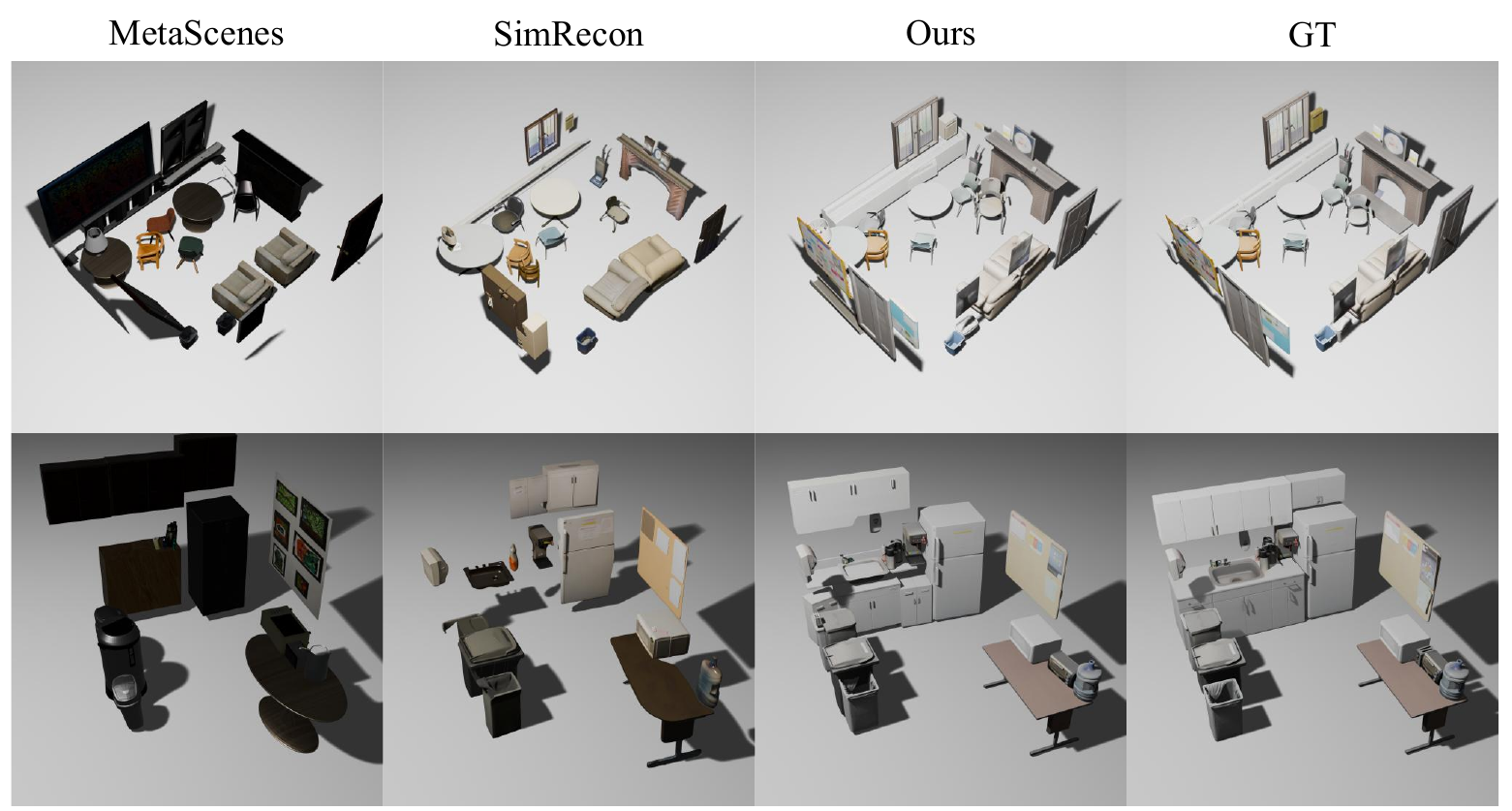}
  \caption{\textbf{Qualitative comparison on the C3DR benchmark.} We present representative qualitative visualizations of the final reconstructed scenes alongside results from MetaScenes~\cite{yu2025metascenes} and SimRecon~\cite{xia2026simreconsimreadycompositionalscene}.} 
  \label{fig:results}
\end{figure}

\begin{table}[t]
  \centering
  \caption{\textbf{Quantitative comparisons on the C3DR benchmark.} We evaluate against MetaScenes~\cite{yu2025metascenes} and SimRecon~\cite{xia2026simreconsimreadycompositionalscene} across textual completeness (Rec, F1), visual quality (PSNR, SSIM~\cite{1284395}, LPIPS~\cite{zhang2018perceptual}, MUSIQ~\cite{ke2021musiq}), and geometric accuracy (CD, F, NC). Our method outperforms all baselines across every evaluation dimension.}
  \label{tab:comparison_VS}
  \begin{tabular}{l|cc|cccc|ccc}
    \toprule
    \multirow{2}{*}{Method} & \multicolumn{2}{c|}{Textual} & \multicolumn{4}{c|}{Visual} & \multicolumn{3}{c}{Geometric} \\
     & Rec$\uparrow$ & F1$\uparrow$& PSNR$\uparrow$ & SSIM$\uparrow$ & LPIPS$\downarrow$ & MUSIQ$\uparrow$ &CD$\downarrow$ & F$\uparrow$ & NC$\uparrow$  \\
    \midrule
    MetaScenes   & \underline{64.29}&  75.10 &  11.63  & 0.746  & 0.243 & 68.18  &  \underline{0.643} &  3.95  & 41.04  \\
    SimRecon        & - &  \underline{80.88} &  \underline{15.77}   & \underline{0.828}  &  \underline{0.206} & \underline{69.49}  & 0.778  & \underline{6.03} & \underline{46.32} \\
    \textbf{Ours} & \textbf{91.43}&  \textbf{90.86} & \textbf{19.45}  & \textbf{0.854}  & \textbf{0.154}  &  \textbf{71.03}  & \textbf{0.614}  &  \textbf{12.91} &  \textbf{48.30} \\
    \bottomrule
  \end{tabular}
\end{table}

\noindent \textbf{Metrics.} Benefiting from the comprehensive annotations provided by our C3DR benchmark, we evaluate reconstruction quality systematically across textual, visual, and spatial dimensions. At the textual level, we measure category completeness via the recall of salient object categories (Rec) and evaluate instance fidelity using the F1-score (F1) to simultaneously capture omissions and spurious predictions. For visual fidelity, we adopt the rendering evaluation setup from ExtraNeRF~\cite{shih2024extranerf} including full-reference metrics such as PSNR, SSIM~\cite{1284395}, and LPIPS~\cite{zhang2018perceptual}, alongside the no-reference perceptual metric MUSIQ~\cite{ke2021musiq}. For spatial accuracy, we follow the standard evaluation protocol from MonoSDF~\cite{yu2022monosdf} by reporting Chamfer Distance (CD), F-Score (F), and Normal Consistency (NC). 

\noindent \textbf{Implementation Details.} All experiments are conducted on a single NVIDIA RTX A6000 GPU. On average, processing one scene takes approximately 27 minutes, with each object requiring about 1.6 minutes. All rendered images used for evaluation are generated in Blender.

\subsection{Results}

\cref{tab:comparison_VS} and \cref{fig:ablation_2_4} present the quantitative and qualitative results for the compositional 3D reconstruction task.

\noindent \textbf{Textual Completeness.} At the textual-level, we evaluate category recall and instance fidelity. Because SimRecon~\cite{xia2026simreconsimreadycompositionalscene} requires category priors, we evaluate category recall solely against MetaScenes~\cite{yu2025metascenes}. Even with access to dataset annotations, MetaScenes~\cite{yu2025metascenes} demonstrates lower category recall than our framework. To fairly assess instance fidelity, we supply SimRecon~\cite{xia2026simreconsimreadycompositionalscene} with the benchmark-defined categories. Under these conditions, our approach consistently achieves higher instance F1-scores by generating complete scenes with minimal spurious or duplicated instances.

\noindent \textbf{Visual Fidelity.} We render images from multiple viewpoints under consistent lighting to evaluate visual quality against ground-truth references. Quantitatively, our method surpasses all baselines across both full-reference and no-reference metrics. Qualitatively, the retrieval-based paradigm of MetaScenes~\cite{yu2025metascenes} heavily restricts its visual fidelity, often resulting in severe color mismatches and shape inconsistencies compared to the actual environments. Our pipeline overcomes these limitations to achieve photorealistic object appearances and strong structural coherence.

\noindent \textbf{Geometric Accuracy.} We compare the reconstructed objects against their ground-truth counterparts to evaluate spatial correctness. SimRecon~\cite{xia2026simreconsimreadycompositionalscene} utilizes 3DGS~\cite{kerbl20233d} to infer object placements, but frequently suffers from floating artifacts that disrupt precise spatial matching. This susceptibility inevitably leads to inaccurate object positions and misaligned layouts. In contrast, our explicitly optimized spatial alignment ensures that the reconstructed scenes remain geometrically accurate and faithfully anchored to the true physical geometry.
\begin{table}[t]
  \centering
  \caption{\textbf{Quantitative ablation study on progressive object discovery.} We compare our approach against Single Frame Merging and Global Video Inference. Reported metrics include recall (Rec) on benchmark-annotated categories, semantic redundancy rate (SRR), the discovery gain ratio (DGR), and total runtime.}
  \label{tab:ablation_1}

\begin{tabular}{l|cccc}
    \toprule
    Method & Rec$\uparrow$ & SRR$\downarrow$ & DGR$\uparrow$ & Time$\downarrow$ \\
    \midrule
    Single Frame Merging & \underline{91.27} & 0.44 & \underline{1.712} & \underline{77s} \\
    Global Video Inference & 77.01 & \textbf{0} & 1.214 & \textbf{18s} \\
    \textbf{Ours} & \textbf{91.43} & \textbf{0} & \textbf{1.715} & 82s \\
    \bottomrule
\end{tabular}
\end{table}

\subsection{Ablation Study}
We conduct ablation studies to validate the effectiveness of our framework by independently evaluating the contribution of each of the five stages.

\begin{figure}[t]
  \centering
  \includegraphics[width=\linewidth]{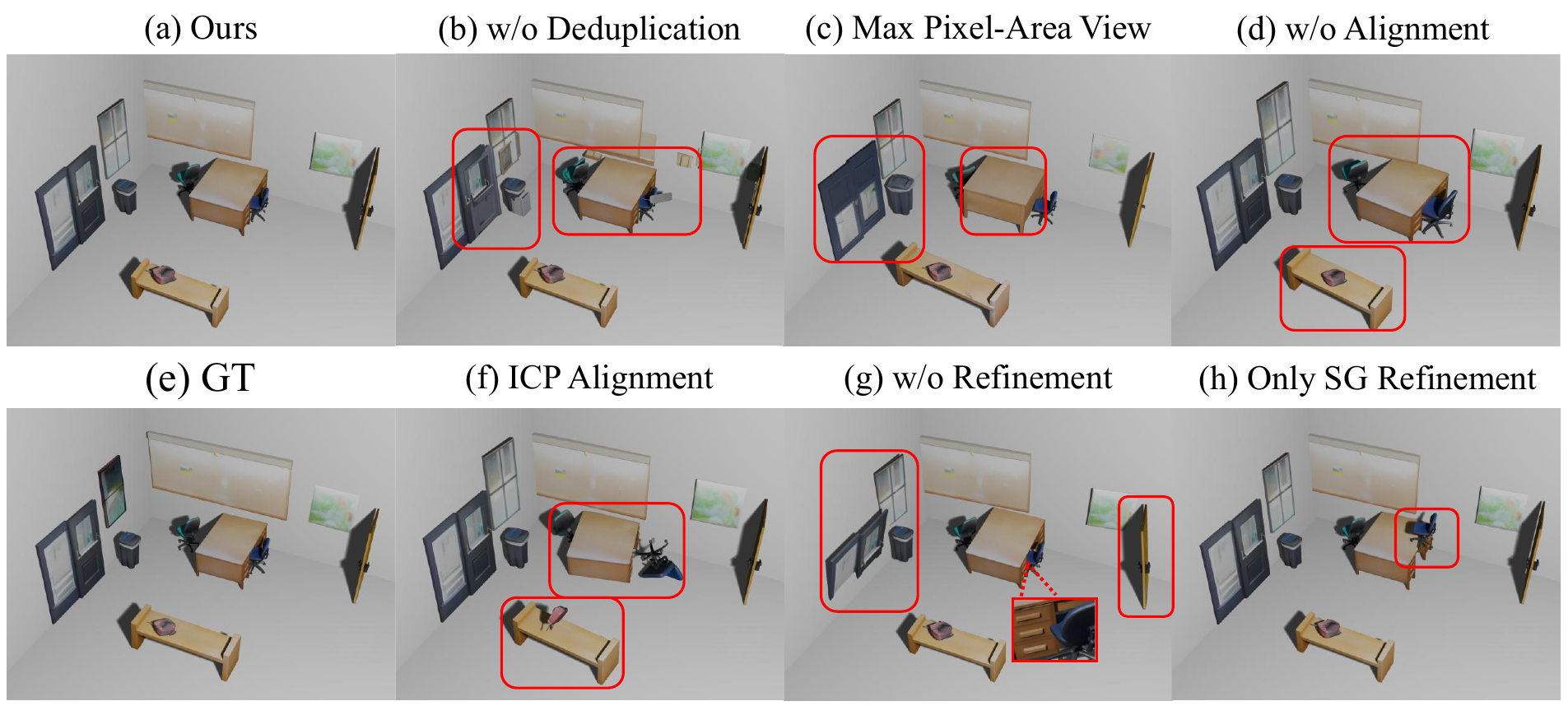}
  \caption{\textbf{Qualitative ablation study on stages 2 to 5 of our pipeline.} Each subfigure shows the scene reconstruction when a specific stage is ablated alongside results from our full method and the ground-truth scene for comparison. Our full pipeline effectively resolves various visual and spatial artifacts. }
  \label{fig:ablation_2_4}
\end{figure}

\begin{table}[t]
\centering
\caption{\textbf{Quantitative ablation study on stages 2 to 5 of our pipeline.} We evaluate across visual quality (PSNR, SSIM~\cite{1284395}, LPIPS~\cite{zhang2018perceptual}, MUSIQ~\cite{ke2021musiq}) and geometric accuracy (CD, F, NC). The results clearly indicate that our full five-stage framework achieves the optimal performance across all evaluated metrics.}
\label{tab:ablation2_5}
\begin{tabular}{l|cccc|ccc}
\toprule
    \multirow{2}{*}{Method} & \multicolumn{4}{c|}{Visual} & \multicolumn{3}{c}{Geometric} \\
 & PSNR$\uparrow$ & SSIM$\uparrow$ & LPIPS$\downarrow$ & MUSIQ$\uparrow$ &CD$\downarrow$ & F$\uparrow$ & NC$\uparrow$  \\
\midrule
w/o Deduplication       & 18.90         & 0.832         & \underline{0.176}  & 69.76          & 0.730       & \underline{12.25}      & 45.38        \\
\midrule
Max Pixel-Area View    & 17.26         & 0.803         & 0.194          & \textbf{71.87}          & \underline{0.729}       & 9.67       & 44.01        \\
\midrule
w/o Alignment       & 16.85         & 0.781         & 0.229          & 70.33          & 0.790       & 12.22      & 43.26        \\
ICP Alignment    & 16.53         & 0.770         & 0.236          & 68.62          & 0.922       & 11.82      & 42.59        \\
\midrule
w/o Refinement       & 16.58         & 0.798         & 0.222          & 68.59          & 0.799       & 9.54       & 44.36        \\
Only SG Refinement      & \underline{19.00} & \underline{0.848}         & 0.163          & 69.73          & 0.744       & 8.55       & \underline{47.34}        \\
\midrule
\textbf{Ours}   & \textbf{19.45} & \textbf{0.854} & \textbf{0.154} & \underline{71.03} & \textbf{0.614} & \textbf{12.91} & \textbf{48.30} \\
\bottomrule
\end{tabular}
\end{table}

\noindent \textbf{Progressive Object Discovery.} We evaluate object category detection using the Qwen3 vision-language model family~\cite{Qwen3-VL} as a representative. We compare our progressive discovery strategy against two baselines including Single Frame Merging baseline, which independently processes single frames and simply merges the predicted categories, and Global Video Inference baseline, which inputs all frames simultaneously in a single inference as shown in \cref{tab:ablation_1}. We report recall on benchmark-annotated categories (Rec) along with total runtime. We additionally compute the semantic redundancy ratio (SRR) and the discovery gain ratio (DGR) to measure comprehensive discovery capabilities. Global Video Inference misses numerous objects, leading to low Rec and DGR scores. Conversely, Single Frame Merging suffers from severe object duplication reflected by a high SRR score. Our method effectively achieves comprehensive object discovery without duplication while maintaining computational efficiency.


\noindent \textbf{Spatial-Guided Visual Deduplication.} Operating without deduplication results in severe object duplication, as clearly illustrated in \cref{fig:ablation_2_4}(b). This specific artifact arises when the same instance is independently detected across multiple non-consecutive video frames.

\noindent \textbf{Optimal-View 3D Asset Generation.} We compare our proposed maximum visible surface area selection against a maximum pixel area baseline. As shown in \cref{fig:ablation_2_4}(c), the pixel baseline introduces severe object artifacts. For large objects like doors and desks, maximum pixel area often simply indicates extreme camera proximity rather than comprehensive visibility. This restricted field of view causes the generation model to produce incomplete and inaccurate textures.

\noindent \textbf{Iterative Visual-Spatial Alignment.} We evaluate our iterative visual-spatial alignment against a configuration without alignment and a standard ICP alignment. As illustrated in \cref{fig:ablation_2_4}(d) and \cref{fig:ablation_2_4}(f), both alternative approaches fail to establish accurate transformation. These failures ultimately manifest as noticeable scaling distortions and severe positional offsets of the reconstructed objects.

\noindent \textbf{Semantic-Aware Scene Refinement.} We evaluate this stage by comparing our method against a configuration without refinement and an only Scene Graph Refinement baseline from SimRecon~\cite{xia2026simreconsimreadycompositionalscene}. As shown in \cref{fig:ablation_2_4}(g), omitting this stage yields physically implausible layouts, including objects floating in mid-air, intersecting the floor, and exhibiting principal axes misaligned with gravity. Our refinement effectively resolves these physical violations. Furthermore, \cref{fig:ablation_2_4}(h) illustrates the only Scene Graph Refinement approach. Relying solely on semantic relations without a robust initial alignment causes severe object misplacements. This demonstrates that semantic relations must serve as a refinement step after a proper alignment, rather than an independent positioning strategy.

\section{Conclusion}

In this paper, we present \textbf{ReplicateAnyScene}, a fully automated and zero-shot framework designed for compositional 3D scene reconstruction from casually captured videos. To overcome the insufficient cross-modal integration and manual dependencies of previous methods, we propose a five-stage cascade pipeline. This architecture effectively extracts and structurally aligns generic priors from vision foundation models across textual, visual, and spatial dimensions, successfully grounding them into coherent 3D representations with strong physical plausibility. Furthermore, to facilitate comprehensive evaluation in this evolving field, we introduced the C3DR benchmark. Extensive experiments demonstrate that ReplicateAnyScene significantly outperforms existing baselines in generating high-quality, structured 3D environments.

\noindent \textbf{Limitations and future work.} Although our framework achieves robust scene replication for indoor environments, it currently encounters difficulties when processing complex natural scenes, because the synthesis of natural objects with highly irregular geometries and the alignment of unbounded natural environments remain well-recognized open problems. In future work, we plan to leverage our pipeline to construct extensive interactive scene datasets, which will significantly facilitate various downstream applications including scene editing agents, video diffusion conditioned on 3D assets, and embodied interaction tasks.

\clearpage  

%
%
\bibliographystyle{splncs04}
\bibliography{main}
\end{document}